\documentclass[lettersize,journal]{IEEEtran}
\usepackage{amsmath,amsfonts}
\usepackage{algorithmic}
\usepackage{algorithm}
\usepackage{array}
\usepackage[caption=false,font=normalsize,labelfont=sf,textfont=sf]{subfig}
\usepackage{textcomp}
\usepackage{stfloats}
\usepackage{url}
\usepackage{verbatim}
\usepackage{graphicx}
\usepackage{cite}
\usepackage{listings}%
\usepackage{pifont}
\usepackage[font=small,labelfont=bf, labelsep=space]{caption}
\begin{document}

\title{Casual Inference via Style Bias Deconfounding for Domain Generalization}

\author{Jiaxi Li, Di Lin,~\IEEEmembership{Member, IEEE}, Hao Chen,~\IEEEmembership{Senior Member, IEEE}, Hongying Liu,~\IEEEmembership{Member, IEEE}, \\ Liang Wan,~\IEEEmembership{Member, IEEE}, and Wei Feng,~\IEEEmembership{Memeber, IEEE}
\thanks{
J. Li, H. Liu and L. Wan are with the Academy of Medical Engineering and Translational Medicine, Tianjin University, Tianjin, China. And L. Wan is also with the College of Intelligence and Computing, Tianjin University, Tianjin 300072, China. E-mail: lijiaxi\_1121@tju.edu.cn; hyliu2009@tju.edu.cn; lwan@tju.edu.cn. (Corresponding author: Liang Wan.)

D. Lin and W. Feng are with the College of Intelligence and Computing, Tianjin University, Tianjin 300072, China. E-mail: ande.lin1988@gmail.com; wfeng@tju.edu.cn.

H. Chen is with the Department of Computer Science and Engineering and the Department of Chemical and Biological Engineering, The Hong Kong University of Science and Technology, Hong Kong, China. E-mail: jhc@cse.ust.hk.}
}

\markboth{IEEE TRANSACTIONS ON PATTERN ANALYSIS AND MACHINE INTELLIGENCE, VOL. X, NO. X, XXX}%
{Shell \MakeLowercase{\textit{et al.}}: A Sample Article Using IEEEtran.cls for IEEE Journals}


\maketitle

\begin{abstract}
Deep neural networks (DNNs) often struggle with out-of-distribution data, limiting their reliability in diverse real-world applications. To address this issue, domain generalization methods have been developed to learn domain-invariant features from single or multiple training domains, enabling generalization to unseen testing domains. However, existing approaches usually overlook the impact of style frequency within the training set. This oversight predisposes models to capture spurious visual correlations caused by style confounding factors, rather than learning truly causal representations, thereby undermining inference reliability.
In this work, we introduce Style Deconfounding Causal Learning (SDCL), a novel causal inference-based framework designed to explicitly address style as a confounding factor.
Our approaches begins with constructing a structural causal model (SCM) tailored to the domain generalization problem and applies a backdoor adjustment strategy to account for style influence. 
Building on this foundation, we design a style-guided expert module (SGEM) to adaptively clusters style distributions during training, capturing the global confounding style. 
Additionally, a back-door causal learning module (BDCL) performs causal interventions during feature extraction, ensuring fair integration of global confounding styles into sample predictions, effectively reducing style bias.
The SDCL framework is highly versatile and can be seamlessly integrated with state-of-the-art data augmentation techniques. 
Extensive experiments across diverse natural and medical image recognition tasks validate its efficacy, demonstrating superior performance in both multi-domain and the more challenging single-domain generalization scenarios.
\end{abstract}

\begin{IEEEkeywords}
Domain Generalization, Casual Inference, Style Bias.
\end{IEEEkeywords}

\section{Introduction}

Deep neural networks (DNNs) have performed remarkably in various visual tasks. However, DNNs typically assume that the training and testing domains are independently and identically distributed, which is not valid in most realistic scenarios~\cite{ben2010theory,vapnik1991principles}. When DNNs are deployed in out-of-distribution (OOD) testing domains, performance may significantly degrade.
To alleviate this problem, domain generalization (DG) methods have been developed to learn domain-invariant features from single or multiple training domains, which can generalize well to unseen test domains. 

\begin{figure}[t]
\centering
\includegraphics[width=\columnwidth]{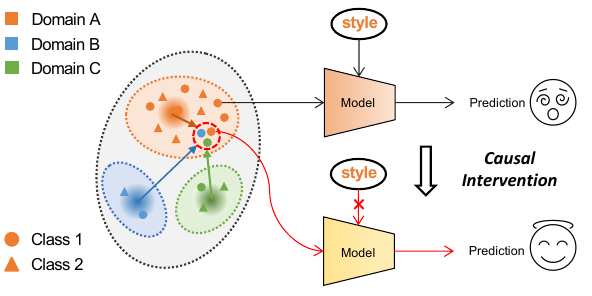}
\caption{A schematic before and after causal intervention. 
Before intervention: The model relies on frequently occurring style types (orange oval) to make predictions.  
After intervention: Different style features from the source domain (green, blue) are fairly incorporated into the prediction of the current sample (orange), enabling the model to consider global styles comprehensively, thus eliminating style bias.}
\label{intro}
\vspace{-4mm}
\end{figure}

Existing DG methods can be categorized as (1) Domain alignment-based: some methods~\cite{zhu2022localized, peng2019moment} attempt to align the features of multiple source domains through feature distribution alignment or feature normalization to learn general representations. However, their  application is limited due to the requirement of data from multiple source domains with different distributions for training. (2) Augmentation-based: most methods~\cite{choi2023progressive, cheng2023adversarial} employ data augmentation to increase the diversity of existing training data, thereby reducing over-fitting and enhancing the model’s generalization capacity. The above methods often incorporate randomness, which may introduce dispersed style complexity and lead to suboptimal generalization. (3) Learning strategies-based: some methods~\cite{wan2022meta, wang2022contrastive} integrate meta-learning and self-supervised learning strategies to learn generalizable representations. These methods typically require domain labels and substantial amounts of data, making them not applicable to a broader range of scenarios. 
Despite showing encouraging performances on OOD data, existing DG methods often focus on modeling statistical dependencies between data $X$ and labels $Y$ while neglecting the impact of style biases. 
As shown in the black line branch of Fig.~\ref{intro}, the model may overfit to the style type represented by the orange points that appear more frequently in the training set, resulting in decreased performance for rare or unseen styles. This over-reliance on specific styles motivates us to develop models that can counteract and mitigate style bias.

Recently, there has been a surge of efforts to integrate causal inference with deep learning for tasks such as image recognition, image captioning, and visual question answering~\cite{mahajan2021domain,wang2022contrastive,lv2022causality,chen2023meta,nguyen2023causal}. This is primarily attributed to its focus on establishing the causal relationship $P(Y|do(X))$ between variables $X$ and $Y$, rather than merely identifying the correlation $P(Y|X)$. Here, the $do(X)$ operator signifies an intervention on $X$, which allows us to eliminate any spurious associations between $X$ and $Y$~\cite{hogan2019causal}. Therefore, to isolate the spurious associations caused by the confounding variable of style, we can employ the well-known causal inference technique of back-door adjustment to estimate $P(Y|do(X))$~\cite{hogan2019causal}. The back-door adjustment formula suggests a feasible approach to control the confounding variable to block spurious associations—by modeling a combination of the input image with other image styles. From a statistical analysis perspective, this can be understood as a ``stratified analysis'' process~\cite{mantel1959statistical}, which involves stratifying the data according to different levels of the confounding variable. This allows for independent analysis within each stratum, ensuring that the confounding variable is constant or similar, effectively controlling its influence.

To incorporate the process of back-door adjustment within a deep learning framework and mitigate the negative impact of confounding styles for causal inference, we propose a novel framework called Style Deconfounding Causal Learning (SDCL). This framework comprises two main modules: the style-guided expert module (SGEM) and the back-door causal learning module (BDCL). SGEM is designed to implement the ``stratification'' process of back-door adjustment. This module adaptively allocates experts based on the style of the samples, achieving style clustering without relying on domain labels. As shown in Fig.~\ref{intro}, points of different colors in the source domain form three cluster centers, which allow for stratifying the global style representation and thus forming confounded set. Building on this, BDCL performs the ``analysis'' process of back-door adjustment. The red branch in Fig.~\ref{intro} illustrates this process. For the current input sample represented by the orange point, this module incorporates the globally stratified style features from the confounded set into the prediction of the current input image through style transfer. This approach compels the model to make decisions across different styles, thereby controlling the influence of confounding styles.

We combined 9 different data augmentation methods, including ABA\cite{cheng2023adversarial}, UDP\cite{guo2023single}, SHADE\cite{zhao2024style}, PASTA\cite{chattopadhyay2023pasta}, etc., and validated them across 7 datasets on 3 image recognition tasks. The experimental results demonstrate that SDCL  
significantly enhances the model's generalization ability in both multi-domain and single-domain scenarios, for instance, an improvement of 1.84 accuracy for single-domain classification (Table~\ref{tab:SDG_Digits})
and 2.43 mIoU for multi-domain segmentation (Table~\ref{tab:MDG_GTAS}).
In summary, the main contributions of this paper are as follows:
\begin{itemize}
\item We introduce a causal-aware domain generalization framework, Style Deconfounding Causal Learning, which mitigates style bias by simulating causal intervention.
\item We propose a style-guided expert module that adaptively clusters the style distributions of the source domain to provide stratified style representations for causal learning.
\item We propose a back-door causal learning module that aggregates input samples and global style representations through causal interventions to combat style bias.
\item SDCL can be integrated with other data augmentation models, showing excellent generalization in multi- and single-domain recognition of natural and medical images.
\end{itemize}

\section{Related work}
\subsection{Data augmentation for domain generalization}
Domain generalization is a promising approach to enhance the generalizability of a model. It aims to train the model using observed data to maintain high performance on a test set with an unknown distribution. Currently, DG methods can be broadly categorized into three dimensions: domain alignment-based approaches~\cite{zhu2022localized, peng2019moment}, generating diverse training samples through data augmentation~\cite{li2021progressive, guo2023single, chattopadhyay2023pasta} and learning strategies to improve generalization ability~\cite{wan2022meta, wang2022contrastive}. The method proposed in this paper builds upon data augmentation algorithms to mitigate the confounding style bias present in the training set. Therefore, this section focuses on data augmentation-based domain generalization methods.

A rich line of domain generalization work employs data augmentation to generate out-of-domain samples that expand the distribution of the source domain. Li et al.~\cite{li2021progressive} proposed a progressive domain expansion network that gradually generates multiple domains by simulating various photometric and geometric transformations through a style shift-based generator. Guo et al.~\cite{guo2023single} synthesized diverse samples by introducing an auxiliary model that maximizes and minimizes the game between the generative model and the auxiliary model in an adversarial manner. Choi et al.~\cite{choi2023progressive} proposed a progressive stochastic convolutional image enhancement by recursively stacking random convolutional layers to create more efficient virtual domains. Chattopadhyay et al.~\cite{chattopadhyay2023pasta} proposed a proportional amplitude spectrum training enhancement strategy, where the amplitude spectrum of the synthesized image is perturbed in the Fourier domain to generate diverse enhanced views. 

The above methods have achieved performance gains in the domain generalization. However, most of these methods cannot control the distribution of the samples they generate, leaving the source domain with scattered data complexity that hinders model learning. To this end, our method introduces causal inference during training to reduce the style bias in the source domain.
    
\subsection{Causal inference for domain generalization}
Causal inference aims to deduce causal relationships between variables from observed data~\cite{pearl2010introduction}. Causality reflects the intrinsic generative mechanism of data, contributing to better out-of-distribution generalization and robustness. Causal inference has been applied to many deep learning research problems, including domain generalization ~\cite{mahajan2021domain,wang2022contrastive,lv2022causality,chen2023meta,nguyen2023causal}.

Many current studies focus on learning domain-invariant semantic representations based on causal relationships, which can be considered direct causes of labeling. For instance, Mahajan et al.~\cite{mahajan2021domain} utilized cross-domain invariance of causal relationships between categories and feature representations for domain generalization. Wang et al.~\cite{wang2022contrastive} focused on the causal invariance of the average causal effect of features on labels, introducing contrast loss to normalize the learning process and enhance the stability of causal predictions across domains. Similarly, Lv et al.~\cite{lv2022causality} extracted causal representations to learn invariant feature representations that mimic the underlying causal factors. On the other hand, some methods aim to identify causality by eliminating spurious correlations caused by confounders. Chen et al.\cite{chen2023meta} used visual attributes, such as brightness, viewpoint, and color, as explicit confounders, analyzing the causes of domain shifts by calculating causal effects and guiding the feature alignment. However, this method requires listing specific confounders and setting hyperparameters to adjust the effects of various factors. Nguyen et al.~\cite{nguyen2023causal} proposed using domain-specific features as implicit confounders. They used style transfer to generate counterfactual images for causal learning to estimate the front-door adjustment formula. However, it must be adjusted based on the adopted style transfer-specific algorithm.

In contrast, our method explicitly targets the fundamental cause of domain shift—style as an explicit confounding factor. By comprehensively understanding the style distribution of the source domain, we mitigate the effects of style bias through the simulation of the back-door adjustment process, thereby uncovering the underlying causal mechanisms. Our work can be viewed as a pioneering attempt to directly address style bias by applying back-door formulas in domain generalization.

\section{Problem Forumlation}\label{Preliminaries}
\subsection{Structural Causal Model of DG}
\begin{figure}
\includegraphics[width=0.48
\textwidth]{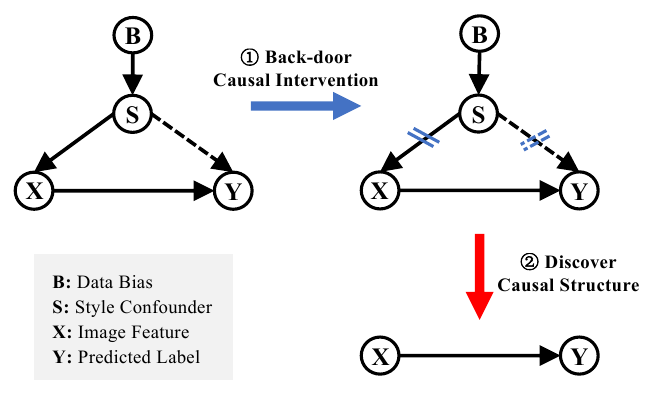}
\centering
\caption{The proposed SCM for DG and the causal inference process. Please refer to the text for detailed explanation.}
\label{SCM}
\end{figure}

We analyze the distribution $P(Y|X)$ from a causal perspective to understand better the limitations of statistical learning in the DG setting. Fig.~\ref{SCM} illustrates a causal graph $G$ that describes the relationship between $X$ and $Y$.
In this graph, along with $X$ and $Y$, two additional random variables, $B$ and $S$, represent the data sampling bias and the domain-specific style, resptively. 
In the practice of DG, selecting a specific type of images can introduce data sampling bias. For example, most autopilot datasets primarily consist of daytime scenes with favorable weather conditions, resulting in the model's tendency to predict outcomes based on domain-specific styles $S$. The relationships and meanings of the connections between the variables in $G$ are detailed below:

$\mathbf{B \rightarrow S \rightarrow X}$: The input image $X$ contains style information that determines the image's appearance. Sampling bias $B$ may lead the model to rely on a specific style for decision-making, which renders style $S$ a confounding factor.

$\mathbf{B \rightarrow S \rightarrow Y}$: Since the sample style $S$ is a confounding factor in the model training process, we must also connect $S$ via a directed path to the prediction $Y$. This ensures that the confounding effects from $S$ to $Y$ are considered.

$\mathbf{X\leftarrow S\rightarrow Y}$: There exists a back-door path where the confounding style $S$ causes a spurious correlation between $X$ and $Y$, i.e., the model is prone to predict the outcome $Y$ based on specific style features.

$\mathbf{X\rightarrow Y}$: Ultimately, the semantic features extracted from the  image $X$ primarily determine the prediction result $Y$.

To train a DG model that learns the true causal effect $X\rightarrow Y$, the model should infer the predicted outcome $Y$ from the semantic feature instead of exploiting spurious correlations induced by confounders $S$. To achieve this, we mitigate the model's bias towards specific styles by applying causal inference to discover the causal structure and intervene with sample features.

\subsection{Causal intervention}
To learn causal relationships between variables, we apply a back-door adjustment strategy~\cite{pearl2010introduction} to eliminate the effects of confounders by modifying the structural causal model and manipulating conditional probabilities. Traditional DG models focus on correlating input images and predicted outcomes by directly learning $P(Y|X)$ without accounting for the confounding factor $S$. In our SCM, the prediction without intervention can be expressed using Bayes rule~\cite{liu2023cross} as follows:
\begin{gather}
P(Y|X) = \sum_s{P(Y|X,s)P(s|X)}.
\end{gather}

We implement a back-door adjustment strategy to eliminate spurious correlations by intervening on $X$ (i.e., artificially fixing the value of $X$). This intervention severs the connection $X \leftarrow S$ and blocks the back-door path $X \leftarrow S \rightarrow Y$. This adjustment can be realized by calculating the intervention distribution using the $do$ operator~\cite{pearl2010introduction}:
\begin{gather}\label{do}
P(Y|do(X=x)) = \sum_s{P(Y|X=x,s)P(s)}.
\end{gather}

From the above equation, we observe that we cut off the back-door path by stratifying the style variable $S$ into various cases $\left \{s \right\}$. This allows us to estimate the average causal effect of $X$ on $Y$. Intuitively, this mimics a physical intervention. For example, if one style predominates in the training data, the model may learn correlations between that style and the predicted outcome while overlooking crucial structural information. By applying causal intervention, we incorporate the styles of other samples into the predictions of the current sample to mitigate this spurious correlation. This necessitates that the model consistently draws inferences from diverse style features. This process enhances the model's ability to extract robust features.

\begin{figure*}[ht]
\centering
\includegraphics[width=1\textwidth]{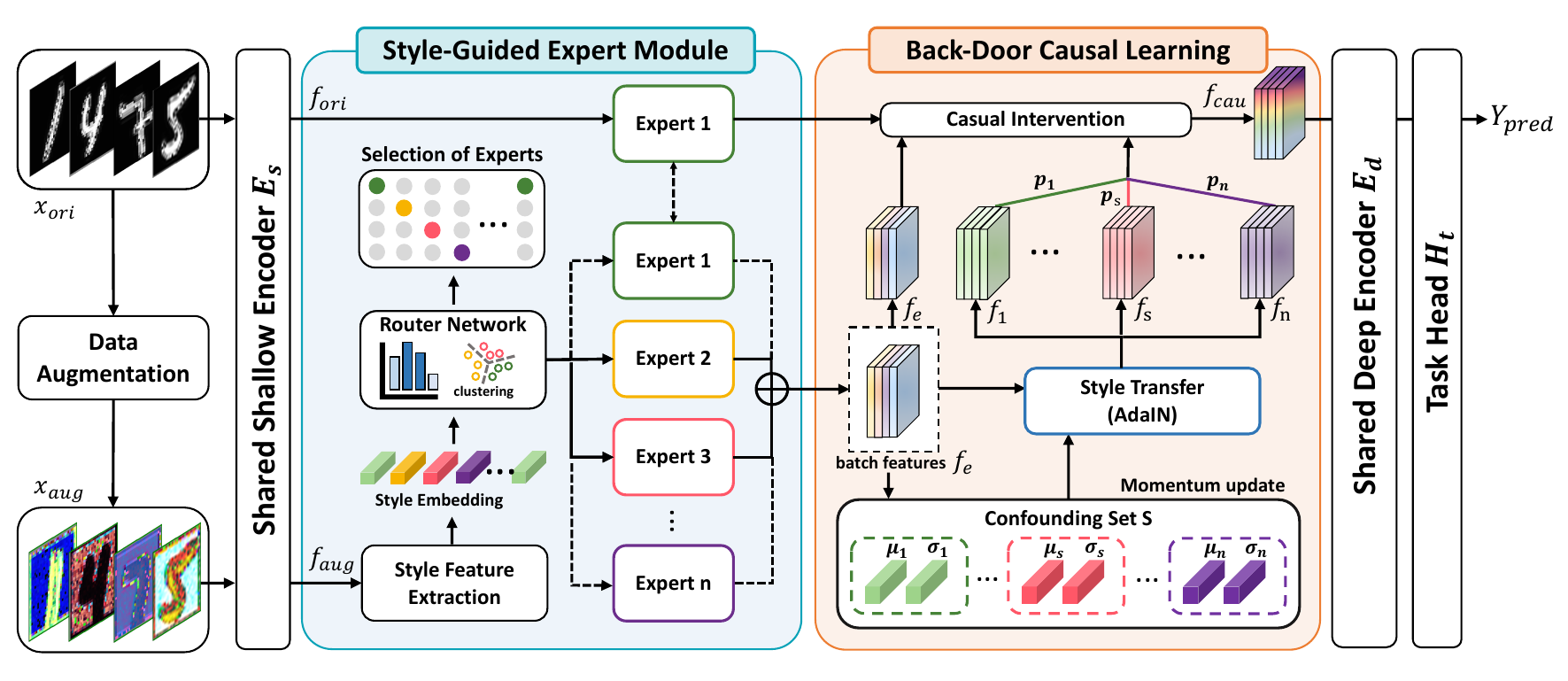}
\caption{Overview of SDCL.
The original and augmented samples are input into SDCL, which first extracts style embeddings from shallow features and inputs them into the SGEM for adaptive expert allocation, constructing a confounded style set. Then, BDCL uses this confounded set to perform style transfer to facilitate sample interventions, obtaining causal features that are input into the subsequent network for classification or segmentation.}\label{method}
\vspace{-4mm}
\end{figure*}
\section{Proposed Method}
In this study, we investigate domain generalization, where a DNN is trained on a single or multiple labeled source domain $D_s = \left \{(x{^s_i}, y_i^s)|_{i=1}^{N_s}\right \}$ drawn from the distribution $P_s$. Here, $x_i^s \in \mathcal{X} \subset \mathbb{R}^X $ represents the $i$-th source sample and $y_i^s \in \mathcal{Y} \subset \mathbb{R}^K $ denotes its corresponding category label. The objective is to utilize the trained DNN to perform well on multiple unseen target domains $D_t = \left \{D_t^z\right \}_{z=1}^Z$ during the testing phase, where each target domain $D_t^z$ is drawn from a distribution $P_t^z$ that differs from $P_s$.
The whole DNN architecture is represented as $F=E \circ H$, where $E:\mathbb{R} ^X \rightarrow \mathbb{R}^D$ denotes the feature extractor and $H:\mathbb{R} ^D \rightarrow \mathbb{R}^K$ is the classifier. We categorize the encoders into shallow encoders $E_s$ and deep encoders $E_d$ to eliminate style confounding. This categorization is based on the fact that shallow layers of DNN usually contain more style information, while deep layers contain much structural-semantic information~\cite{zeiler2014visualizing}. Thus, we can perform the hierarchical style feature and sample intervention on the extracted shallow features.

Our proposed SDCL consists of two modules to enable the model to simulate the causal inference process. The SGEM first constructs the confounding set by using sample styles as conditions and adaptively assigning experts to achieve clustering of the different style sub-distributions. Next, BDCL mitigates the effects of confounding styles by fusing features from samples of different styles to estimate a back-door adjustment formula. An overview is shown in Fig.~\ref{method}.

\subsection{Style-Guided Expert Module}\label{SGEM}
In the context of DG, the source domain data is the foundation for model training. 
Leveraging diverse data can introduce complexity due to the dispersion of style distributions present within the source domain. 
To simulate the modeling process described in Eq.~\ref{do} and address the style bias inherent in these distributions, we propose a style-guided expert module to identify and hierarchize the confounding factors $S$. Our method employs the Mixture of Experts (MoE) framework~\cite{shazeer2017outrageously}, which efficiently creates a clustering structure for style distributions by using sample styles as routing conditions, thus better managing the style variability of the training data.

Specifically, the SGEM consists of a set of expert networks $\left \{E_1, E_2, \dots, E_n\right\}$ and a routing network $R$. Each expert outputs $E_k(X)$ for $k = 1, 2, ..., n$, where the overall output is a weighted sum of the expert outputs determined by the routing network. Input a batch of samples $X \in \mathbb{R}^{B \times C \times H \times W}$, where $B$, $C$, $H$, and $W$ denote the batch size, number of channels, height, and width of the samples, respectively. To facilitate targeted expert selection based on sample styles, we first fix the input into expert $E_1$ for the original sample features $f_{ori}$ that are relatively consistent in style. We then extract style statistics~\cite{huang2017arbitrary} (i.e., mean and standard deviation across channels) from the augmented features $f_{aug}$. These statistics are combined into a style embedding $Z$, defined as:
\begin{gather}\label{style_em}
Z = concat\left [\mu(x),\sigma(x) \right ], \\
\mu(x) = \frac{1}{HW} \sum_{h=1}^{H}\sum_{w=1}^{W} x, \\
\sigma(x) = \sqrt{\frac{1}{HW} \sum_{h=1}^{H}\sum_{w=1}^{W}(x - \mu(x))^2},
\end{gather}
After obtaining the style embedding $Z$, it is fed into the router network $R$. To improve efficiency and optimize model parameters, we employ a Topk strategy to activate only the Top K experts with the highest weights, thereby enabling the model to effectively leverage the specializations of different experts when dealing with complex styles. Consequently, the output $f_e$ of the expert network is represented as follows:
\begin{gather}\label{out}
f_e= \sum\limits_{i=1}^k R(Z) \cdot E_i(x),\\
R(Z) = \mathrm{Softmax}(\mathrm{Topk(\mathrm{Softmax}}(R(Z), \mathrm{k})).
\end{gather}

Then, we can construct the style confounding set based on the expert network approximation. By analyzing the output weights from the $R$ network, we can identify which style distributions each expert is best equipped to handle. Consequently, we define the style confounding set as $S=\left [(\mu_1,\sigma_1),\dots, (\mu_s,\sigma_s),\dots, (\mu_n,\sigma_n) \right]$, and a pair of means and standard deviations that represent the distribution of styles of samples that select the same expert, then $(\mu_s, \sigma_s) $ can be computed as~\cite{hong2021stylemix}:
\begin{gather}
\mu_s = \frac{1}{N_b}\sum_{i=1}^B\mathbb{I}[\mathrm{argmax}(R(Z))=s]\,\mu(f_e^i), \\
\sigma_s = \frac{1}{N_b}\sum_{i=1}^B\mathbb{I}[\mathrm{argmax}(R(Z))=s]\,\sigma(f_e^i),
\end{gather}
where $\mathbb{I}(\cdot)$ is the indicator function, $N_b$ denotes the number of selected experts $s$ in all the samples in a batch, and $\mu(f_e^i)$ and $\sigma(f_e^i)$ denote the mean and the standard deviation of the sample features of the selected experts $s$ in the channel dimension, respectively. As training progresses, we refine these estimates using a momentum mechanism, achieving a hierarchical representation of style distributions:
\begin{gather}
\mu_s = (1 - \tau) \mu^t_s + \tau \mu^{t-1}_s, \sigma_s = (1 - \tau) \sigma^t_s + \tau \sigma^{t-1}_s,
\end{gather}
where $\tau$ is the momentum update coefficient, $t$ is the training step, and $\tau$ is uniformly set to 0.9 in the experiment.

Additionally, to ensure that all experts are actively trained and not just a limited few, we introduce a regularization term $L_{reg}$~\cite{shazeer2017outrageously} to balance the allocation of expert resources:
\begin{gather}\label{reg}
\mathcal{L}_{reg}=CV\left(\sum_{x\in X} R\left(x\right)\right)^2,
\end{gather}
where $CV$ denotes the Coefficient of Variation~\cite{abdi2010coefficient}. This strategy encourages each expert to specialize in addressing a particular style distribution, ultimately facilitating the approximation of the style confounding set.

\subsection{Back Door Casual Learning based on Style Bias}
To mitigate style bias and reveal potential causal relationships, we propose a back-door causal learning strategy informed by causal inference principles discussed in Section~\ref{Preliminaries}.

Our goal is to learn deconfounded features for each sample by estimating $P(Y|do(X))$ as outlined in Eq.~\ref{do}. The critical aspect of this estimation is computing $P(Y|X,s)$. Here, $P(s)$ is the a prior probability, and to fairly incorporate the different styles into the prediction of the current sample, we set $P(s)$ to a uniform value of $1/n$, following the established work~\cite{zhang2020causal}. The prediction of $P(Y|X,s)$ is usually computed by softmax, i.e., $P(Y|X,s) = \mathrm{Softmax}(f(Y|X,s))$, where $f(Y|X,s)$ represents the original features of sample $X$ and the fusion result with the features of style $s$. Specifically, we can reformulate Eq.~\ref{do} as:
\begin{align}
P(Y|do(X)) &= \sum_s{P(Y|X,s)P(s)} \nonumber \\
           &= \mathbb{E}_s\left[(P(Y|X,s)\right] \\ \nonumber 
           &= \mathbb{E}_s\left[\mathrm{Softmax}(f(X, s)\right],
\end{align}
Next, we approximate $\mathbb{E}_s\left[\mathrm{Softmax}(f(X, s)\right]$ using the normalized weighted geometric mean (NWGM)~\cite{xu2015show}, which enables back-door adjustment at the feature level:
\begin{align}\label{NWGM}
\mathbb{E}_s\left[\mathrm{Softmax}(f(X, s)\right] &\approx \mathrm{NWGM}[\mathrm{Softmax}(f(X, s))] \nonumber \\
           &= \mathrm{Softmax}(\mathbb{E}_s(f(X, s))\\
    \nonumber   &= \mathrm{Softmax}(\sum_{s}f(X,s) P(s)).
\end{align}

To eliminate the back-door path $X \rightarrow S \leftarrow Y$ and ensure that each $s$ incorporates causality-aware factors in predictions (as shown in Fig.~\ref{SCM}), we apply back-door causal interventions by merging current features with those representing different styles. First, we utilize the Adaptive Instance Normalization (AdaIN)~\cite{huang2017arbitrary} operator to create style-specific features $f_s$ from the distributions in the confounding set $S$:
\begin{gather}
f_s = (\sigma_s \odot \epsilon)\, \frac{f_e\ -\ \mu(f_e)}{\sigma(f_e)} + \mu_s, \quad \epsilon \sim  N(0, I_d),
\end{gather}
in this expression, $\epsilon$ is a noise perturbation added within a controlled range to enhance the robustness of the model. 
We further simulate the process in Eq.~\ref{NWGM} using feature fusion, i.e., incorporating the confounding style features into the features of the input image, whereby $f(X, s)$ is defined as $(\alpha f_e + (1 - \alpha) f_s)$. The causal features generated within Softmax in Eq.~\ref{NWGM} can be represented as:
\begin{gather}\label{fusion}
f_{cau} = \alpha f_e + (1 - \alpha) \sum_{s}f_s P(s),
\end{gather}
where $\alpha$ is the feature fusion coefficient. Each style distribution in $S$ is incorporated into the model training phase utilizing Eq.~\ref{fusion}, which  represents a weighted sum of occurrences stemming from style confounders within the source domain.

Finally, the intervened features $f_{cau}$ are input into the remaining task network, where we employ task loss to enforce consistency regularization~\cite{guo2023single}. The overall training objective of the network is formulated as follows:
\begin{gather}
\mathcal{L} = \mathcal{L}_{Task}(f_{ori},y) + \mathcal{L}_{Task}(f_{aug},y) + \lambda \mathcal{L}_{reg},
\end{gather}
where $\lambda$ is the weighting factor of the regularization loss and is set to 1.0 in the experiment.

During inference, the test samples are input into the shallow encoder $E_s$, then adaptively assigned appropriate expert processing through SGEM, and finally input into the deep encoder $E_d$ and task head $H_t$ for classification or segmentation.
\section{Experiments}
We assess the effectiveness of the SDCL framework for DG through a series of empirical evaluations across various tasks, including the classification of natural images and the segmentation of both natural and medical images.

\subsection{Image Classification}
We evaluate our method against several state-of-the-art methods, including the vanilla strategy (ERM\cite{vapnik1999overview}), methods that learn domain-invariant features (
MetaCNN\cite{wan2022meta}, VL2V\cite{addepalli2024leveraging}), methods that incorporate causality (CIRL\cite{lv2022causality}, MetaCAU\cite{chen2023meta}, LRDG~\cite{ding2022domain}, XDED~\cite{huang2020self}), and various data augmentation strategies (L2D\cite{L2D}, GeoTexAug\cite{liu2022geometric}, RSC\cite{lee2022cross}, NormAUG~\cite{qi2024normaug}). Our SDCL excels in fusing features from different sub-distributions while utilizing image augmentation to facilitate causal interventions in the data. As a result, SDCL can be effectively integrated with data augmentation-based methods (ABA\cite{cheng2023adversarial}, UDP\cite{guo2023single}, AdvST~\cite{zheng2024advst}, SHADE\cite{zhao2024style}), leading to further performance enhancements.

\subsubsection{Datasets and Implementation Details}
We conducted experiments on two benchmark datasets for DG. Digits contains five domains: MNIST\cite{MNIST}(MN), SVHN\cite{SVHN}(SV), MNIST-M\cite{MNISTM-SYN}(MM), USPS\cite{USPS}(US), and SYN\cite{MNISTM-SYN} (SY). PACS\cite{li2017deeper} contains four domains: Photo(P), Art Painting(A), Cartoon(C), and Sketch(S).
For a fair comparison, we follow the existing dataset partitioning and validation approach for domain generalization methods. We used ConvNet~\cite{guo2023single} as the backbone for the Digits dataset with a batch size of 256 and the structure $conv1$-$bn$-$conv1$-$bn$. For the PACS dataset, we employed ResNet18~\cite{he2016deep} with a batch size of 128 and an expert network structured at layer 1.
Each model was trained for 300 epochs, evaluating performance using classification accuracy. 
We optimized using the Adam optimizer~\cite{kingma2014adam} at a learning rate of $10^{-3}$. Hyper-parameters were set as follows: total number of experts $n=6$, Topk experts $k=4$, and feature fusion coefficient $\alpha=0.7$. The task loss is cross-entropy.

\begin{table}[t]
    \centering
    \caption{\textit{Singel-domain DG classification} accuracies (\%) on \textit{Digits} with ConvNet as the backbone. The model is trained on \textit{MNIST}, and evaluated on \textit{SVHN, SYN, MNIST-M,} and \textit{USPS.} \dag indicates the result re-implemented by the officially code.}
    \label{tab:SDG_Digits}
    \footnotesize
    \renewcommand\arraystretch{1.6} 
    \setlength{\tabcolsep}{1.6mm}
    {
    \begin{tabular}{c|c|cccc|c}
    \hline
      Methods
      &Venue
      &SV &SY &MM &US &Avg\\
      \hline
      {ERM} & &27.83 &52.72 &39.65 &76.94 &49.29\\
      \hline
      {L2D}  &ICCV 21       &62.86 &87.30 &63.72 &83.97 &74.46\\
      {MetaCNN} &CVPR 22    &66.50 &88.27 &70.66 &89.64 &78.76\\
      {CIRL} &CVPR 22       &64.32 &80.23 &74.32 &87.49 &76.59\\
      {MetaCAU} &CVPR 23    &69.94 &78.47 &78.34 &88.54 &78.82\\
      {VL2V} &CVPR 24  &70.05 &79.94 &79.88 &87.04 &79.22\\
      \hline
      {ABA} &ICCV 23        &65.60 &82.25 &77.05 &91.67 &79.14\\
      {ABA w/SDCL} & /        &66.55 &81.44 &78.62 &93.07 &\textbf{79.92}\\
      \hline
      {UDP} &MM 23          &72.38 &79.70 &81.73 &96.26 &82.52\\
      {UDP w/SDCL} & /        &73.03 &86.57 &83.47 &94.36 &\textbf{84.36}\\
      \hline
      {AdvST} &AAAI 24      &67.50 &79.80 &78.10 &94.80 &80.10\\
      {AdvST w/SDCL} & /      &70.05 &80.45 &80.30 &93.70 &\textbf{81.12}\\
      \hline
      {SHADE$^\dag$} &IJCV 24      &70.02 &80.65 &79.84 &94.32 &81.20\\
      {SHADE w/SDCL} & /      &72.38 &81.06 &79.96 &94.89 &\textbf{82.07}\\
      \hline
    \end{tabular}}        
    \end{table}

\begin{table}[h]
    \centering
    \caption{\textit{Singel-domain DG classification} accuracies (\%) on \textit{PACS} with ResNet18 as the backbone. One domain (name in column) is the source domain, and the other three are the target domains. \dag indicates the result re-implemented by the officially code.}
    \label{tab:SDG_PACS}
    \footnotesize
    \renewcommand\arraystretch{1.6} 
    \setlength{\tabcolsep}{1.6mm}{
    \begin{tabular}{c|c|cccc|c}
    \hline
      Methods
      &Venue
      &P &A &C &S &Avg\\
      \hline
      {ERM} &                 &42.20 &70.90 &76.50 &53.10 &60.70\\
      \hline
      {L2D}  &ICCV 21       &52.29 &76.91 &77.88 &53.66 &65.18\\
      {GeoTexAug} &CVPR 22  &49.07 &72.07 &78.70 &59.97 &65.00\\
      {CIRL} &CVPR 22       &53.74 &69.97 &77.42 &58.95 &65.02\\
      {MetaCAU} &CVPR 23    &59.60 &77.13 &80.14 &62.55 &69.86\\
      {VL2V} &CVPR 24  &60.94 &74.91 &80.74 &65.39 &70.49\\
      \hline
      {ABA} &ICCV 23        &58.86 &75.34 &77.49 &53.76 &66.36\\
      {ABA w/SDCL} & /       &60.30 &76.40 &80.34 &54.21 &\textbf{67.81}\\
      \hline
      {UDP} &MM 23          &57.22 &78.61 &73.93 &55.73 &66.37\\
      {UDP w/SDCL} & /       &59.08 &75.40 &74.96 &61.90 &\textbf{67.83}\\
      \hline
      {AdvST$^\dag$} &AAAI 24      &64.10 &78.29 &75.54 &54.37 &68.07\\
      {AdvST w/SDCL} & /      &68.04 &79.49 &78.45 &58.04 &\textbf{71.00}\\
      \hline
      {SHADE$^\dag$} &IJCV 24      &63.90 &79.14 &74.82 &56.03 &68.47\\
      {SHADE w/SDCL} & /      &66.32 &80.74 &76.18 &58.36 &\textbf{70.40}\\
      \hline
    \end{tabular}}
\end{table}

\begin{table}[t]
    \centering
    \caption{\textit{Multi-domain DG classification} leave-one-domain-out results accuracies (\%) on \textit{PACS} with ResNet18 as the backbone. \dag indicates the result re-implemented by the officially code.}
    \label{tab:MDG_PACS}
    \footnotesize
    \renewcommand\arraystretch{1.6} 
    \setlength{\tabcolsep}{1.6mm}{
    \begin{tabular}{c|c|cccc|c}
    \hline
      Methods
      &Venue
      &P &A &C &S &Avg\\
      \hline
      {ERM} &               &87.84 &67.52 &76.66 &76.88 &77.22\\
      \hline
      {RSC}   &ECCV 20      &83.43 &80.31 &95.99 &80.85 &85.15\\
      {LRDG}  &NIPS 22      &81.88 &80.20 &95.21 &84.65 &85.48\\
      {CIRL}  &CVPR 22      &86.08 &80.59 &95.93 &82.67 &86.32\\
      {XDED}  &ECCV 22      &85.60 &84.20 &96.50 &79.10 &86.35\\
      {NormAUG} &TIP 24     &85.60 &81.85 &95.70 &85.00 &87.04\\
      \hline
      {UDP$^\dag$} &MM 23          &85.71 &81.58 &94.22 &83.69 &86.30\\
      {UDP w/SDCL} & /      &88.43 &85.40 &94.05 &86.48 &\textbf{88.59}\\
      \hline
      {SHADE$^\dag$} &IJCV 24      &81.90 &81.43 &95.03 &82.88 &85.31\\
      {SHADE w/SDCL} & /    &83.93 &82.26 &94.85 &86.79 &\textbf{86.95}\\
      \hline
    \end{tabular}}
    \vspace{-4mm}
\end{table}

\subsubsection{Experiment Results}
We began our evaluation with the Digits dataset, the results of which are presented in Table~\ref{tab:SDG_Digits}. Most methods outperformed the baseline ERM, and while they achieved commendable results, our SDCL further enhanced their performance. Specifically, SDCL raised accuracy of UDP from 82.52\% to 84.36\%, securing the top rank among all evaluated methods.

Tables~\ref{tab:SDG_PACS} and~\ref{tab:MDG_PACS} summarise the performance of single-domain and multi-domain generalization on the PACS dataset, respectively. PACS is regarded as more challenging due to significant style variations among its domains, which allows existing data augmentation methods that focus on these differences to yield substantial performance improvements. In single-domain generalization, similar to our findings with the Digits dataset, we observed consistent improvements when integrating data augmentation techniques with SDCL. For instance, SDCL improved the accuracy of AdvST from 68.07\% to 71.00\%. SDCL is also competitive in multi-domain generalization and improves by 1.55\% over the SOTA NormAUG while outperforming UDP and SHADE by 2.29\% and 1.64\%, respectively. Experimental results demonstrate the potential of SDCL for classification.

\subsection{Natural Image Segmentation}
Image segmentation is a crucial skill in autonomous driving, but models in this area often experience significant performance decline due to variations in road scenes. In this section, we conduct experiments on a natural image segmentation task, comparing our approach with several leading methods. These include the vanilla strategy (ERM\cite{vapnik1999overview}), techniques that learn domain-invariant features (IBN-Net\cite{pan2018two}, RobustNet\cite{choi2021robustnet}, WildNet\cite{lee2022wildnet}, SAN-SAW\cite{peng2022semantic}, MLDG\cite{li2018learning}, AdvStyle\cite{zhong2022adversarial}), and methods that focus on generated data augmentation (SHADE\cite{zhao2024style}, TLDR\cite{kim2023texture}, PASTA\cite{chattopadhyay2023pasta}). Consistent with the classification task above, we added our approach to several data augmentation methods to further improve performance.

\subsubsection{Datasets and Implementation Details}
For our experiments, we utilized the GTAV~\cite{richter2016playing} and SYNTHIA~\cite{ros2016synthia} as our synthetic source, which comprises 19 annotated classes that are compatible with the target datasets Cityscape~\cite{cordts2016cityscapes}(C), BDD100K~\cite{yu2020bdd100k}(B), and Mapillary~\cite{neuhold2017mapillary}(M). 
We trained our model using the training splits from the synthetic dataset and evaluated it on the validated splits of the real target datasets, reporting performance based on mean Intersection over Union (mIoU). For a fair comparison, our experimental setup is consistent with the comparison method, both using the DeepLabV3+~\cite{chen2018encoder} with ResNet101~\cite{he2016deep} and ResNet50~\cite{he2016deep} as the backbone for the segmentation network. We optimized the model with a Stochastic Gradient Descent (SGD) optimizer, using a momentum of 0.9, a batch size of 8, and a learning rate of 5e-4. All other parameters remained consistent with those used in the classification tasks. The task loss is the cross-entropy loss for semantic segmentation.

\subsubsection{Experiment Results}
The average results for real-world datasets under single-domain and multi-domain generalization are given in Tables~\ref{tab:SDG_GTAV} and~\ref{tab:MDG_GTAS}, respectively. The results show that our method consistently improves performance by combining the three existing data enhancement methods. In single-domain generalization, our SDCL method enhances the TLDR approach by an average of 1.47\%, achieving state-of-the-art overall performance. We also achieved an average improvement of 1.6\% over other benchmark methods for the more challenging BDD100K domain. In multi-domain generalization, SDCL outperforms the baseline model PASTA by 2.43\% while achieving an optimal performance of 46.90\%. These advancements are attributed to the causal features learned during training with SDCL, highlighting the benefits of employing a causal inference framework.

\begin{table}[t] 
    \centering
    \caption{\textit{Single-domain DG segmentation} mIoU (\%) on real datasets with ResNet101 as the backbone. \textit{GTAV} is used as the source domain, and the other datasets are used as the target domains.}
    \label{tab:SDG_GTAV}
    \footnotesize
    \renewcommand\arraystretch{1.6} 
    \setlength{\tabcolsep}{2mm}{
    \begin{tabular}{c|c|ccc|c}
    \hline
      Methods
      &Venue
      &C &B &M &Avg\\
      \hline
      {ERM} &                  &35.16 &29.71 &31.29 &32.05\\
      \hline
      {IBN-Net}   &ECCV 18   &37.37 &34.21 &36.81 &36.13\\
      {RobustNet} &CVPR 21   &37.20 &33.36 &35.57 &35.38\\
      {WildNet}   &CVPR 22   &45.79 &41.73 &47.08 &44.74\\
      {SAN-SAW}   &CVPR 22   &45.33 &41.18 &40.77 &42.43\\
      \hline
      {TLDR}    &ICCV 23     &47.58 &44.88 &48.80 &47.00\\
      {TLDR w/SDCL}  & /       &49.93 &46.56 &48.94 &\textbf{48.47}\\
      \hline
      {PASTA}    &ICCV 23    &45.33 &42.32 &48.60 &45.42\\
      {PASTA w/SDCL}  & /      &46.84 &43.91 &49.19 &\textbf{46.64}\\
      \hline
      {SHADE}    &IJCV 24    &46.66 &43.66 &45.50 &45.27\\
      {SHADE w/SDCL}  & /      &47.58 &44.95 &45.82 &\textbf{46.11}\\
      \hline
    \end{tabular}}        
\end{table}

\begin{table}[t] 
    \centering
    \caption{\textit{Multi-domain DG segmentation} mIoU (\%) with ResNet50 as the backbone. \textit{GTAV} and \textit{SYNTHIA} are used as the source domain. \dag indicates the result re-implemented by the officially code.}
    \label{tab:MDG_GTAS}
    \footnotesize
    \renewcommand\arraystretch{1.6} 
    \setlength{\tabcolsep}{2mm}{
    \begin{tabular}{c|c|ccc|c}
    \hline
      Methods
      &Venue
      &C &B &M &Avg\\
      \hline
      {ERM} &                  &35.46 &25.09 &31.94 &30.83\\
      \hline
      {IBN-Net}   &ECCV 18     &35.55 &32.18 &38.09 &35.27\\
      {MLDG}      &AAAI 18     &38.84 &31.95 &35.60 &35.46\\
      {RobustNet} &CVPR 21     &37.69 &34.09 &38.49 &36.76\\
      {AdvStyle}  &NeurIPS 22  &39.29 &39.26 &41.14 &39.90\\
      \hline
      {TLDR$^\dag$}    &ICCV 23       &41.31 &39.85 &45.99 &42.38\\
      {TLDR w/SDCL}  & /       &44.76 &41.54 &47.09 &\textbf{44.40}\\
      \hline
      {PASTA$^\dag$}    &ICCV 23      &43.12 &42.08 &45.82 &43.67\\
      {PASTA w/SDCL}  & /      &44.73 &45.85 &47.98 &\textbf{46.10}\\
      \hline
      {SHADE}    &IJCV 24      &47.43 &40.30 &47.60 &45.11\\
      {SHADE w/SDCL}  & /      &46.93 &45.60 &48.17 &\textbf{46.90}\\
      \hline
    \end{tabular}}        
\end{table}

\subsection{Medical Image Segmentation}
The above experiments primarily focus on DG for natural images, but medical image segmentation models often struggle with performance due to variations across devices and modalities. We apply our approach to medical images and compare its effectiveness against several state-of-the-art methods, including the techniques for learning domain-invariant features (VFMNet\cite{liu2022vmfnet}, CDDSA\cite{gu2023cddsa}, LRSR\cite{chen2024learning}), and feature augmentation methods (MaxStyle\cite{chen2022maxstyle}, Trid\cite{chen2023treasure}). Additionally, we integrate our method with medical image augmentation methods (GIN\cite{ouyang2022causality}, SLAug~\cite{su2023rethinking}) and natural image augmentation methods (RandConv\cite{xurobust}, UDP\cite{guo2023single}) to evaluate its effectiveness.

\subsubsection{Datasets and Implementation Details}
We evaluate two tasks: cross-modal abdominal segmentation (CMAD) and cross-device cardiac segmentation (CDCS). CMAD uses datasets CHAOS\cite{kavur2021chaos} (20 MRI cases) and BTCV\cite{landman2015miccai} (30 CT cases), while CDCS employs the M\&Ms dataset~\cite{campello2021multi}, which includes domains A, B, C, and D. We used the average Dice score for evaluation. We trained on one domain and tested across the others, using the average Dice score for evaluation. The segmentation network is DeepLab-V3+~\cite{chen2018encoder} with ResNet50~\cite{he2016deep} as the backbone, optimized with Adam at a learning rate of 3e-4 and a batch size of 20. The model was trained for 2000 epochs for abdominal datasets and 200 epochs for cardiac datasets. The expert network structure consisted of layer 1, with parameters $n = 4$, $k = 2$, and $\alpha = 0.7$. The Dice loss was utilized as the loss function for the task. The difference in the number of experts $n$ is due to medical images being grayscale, which results in less style variation and fewer experts needed for clustering.

\begin{table*}[t]
  \centering
  \caption{\textit{Single-domain DG segmentation} Dice score (\%) on cross-device cardiac dataset and cross-modality abdominal dataset with ResNet50 as backbone.One domain is used as the source domain, and the other domains are used as the target domains. \dag indicates the result re-implemented by the officially code.} 
  \label{tab:SDG_Med}
  \footnotesize
  \renewcommand\arraystretch{1.5} 
  \setlength{\tabcolsep}{2mm}{
  \begin{tabular}{c|c|cccc|c|cc|c}
  \hline
  Methods &Venue
  &A $\rightarrow$ BCD &B $\rightarrow$ ACD &C $\rightarrow$ ABD &D $\rightarrow$ ABC &Avg &MR $\rightarrow$ CT &CT $\rightarrow$ MR &Avg\\
  \hline
  {MaxStyle$^\dag$} &MICCAI 22
  &73.33 &75.79 &69.22 &65.73 &71.01
  &64.81 &76.53 &70.67 \\
  {Trid$^\dag$} &MICCAI 23
  &83.14 &84.24 &82.24 &82.00 &82.90
  &74.68 &74.75 &74.71 \\
  {VFMNet$^\dag$} &MICCAI 22 &76.03 &74.35 &80.45 &70.04 &75.21
  &62.45 &79.04 &70.74 \\
  {CDDSA$^\dag$} &MIA 23 &78.94 &82.42 &80.94 &75.93 &79.55
  &72.03 &75.76 &73.89 \\
  {LRSR$^\dag$} &TMI 24 &79.95 &84.37 &81.12 &76.84 &80.57
  &74.73 &78.41 &76.57 \\
  \hline
  {RandConv$^\dag$} &ICLR 21
  &72.26 &76.42 &72.85 &68.70 &72.55
  &76.05 &76.34 &76.19 \\
  {RandConv w/SDCL} & /
  &73.84 &77.95 &72.41 &70.84 &\textbf{73.76}
  &79.05 &78.93 &\textbf{78.99} \\
  \hline
  {GIN$^\dag$} &TMI 22
  &73.81 &79.31 &70.01 &70.59 &73.43
  &78.52 &86.31 &82.41 \\
  {GIN w/SDCL} &/
  &72.93 &79.91 &72.56 &72.08 &\textbf{74.37}
  &78.56 &86.95 &\textbf{82.75} \\
  \hline
  {UDP$^\dag$} &MM 23
  &81.34 &82.76 &80.66 &80.94 &81.42
  &82.42 &86.11 &84.26  \\
  {UDP w/SDCL} &/
  &82.34 &84.82 &83.70 &81.09 &\textbf{83.19}
  &83.13 &86.83 &\textbf{84.98} \\
  \hline
  {SLAug$^\dag$} &AAAI 23
  &83.10 &84.25 &80.57 &81.58 &82.37
  &82.42 &86.36 &84.39 \\
  {SLAug w/SDCL} &/
  &82.88 &84.02 &83.35 &81.40 &\textbf{82.91}
  &83.53 &87.47 &\textbf{85.50} \\
  \hline
  \end{tabular}}    
\end{table*}

\begin{table}[h]
  \centering
  \caption{\textit{Multi-domain DG segmentation} leave-one-domain-out Dice score (\%) on \textit{CDCS} with ResNet50 as the backbone. \dag indicates the result re-implemented by the officially code.} 
  \label{tab:MDG_Med}
  \footnotesize
  \renewcommand\arraystretch{1.6} 
  \setlength{\tabcolsep}{1.6mm}{
  \begin{tabular}{c|c|cccc|c}
  \hline
  Methods &Venue
  &A &B &C &D &Avg\\
  \hline
  {Trid$^\dag$} & MICCAI 23
  &83.03 &85.06 &87.19 &87.22 &85.62\\
  {LRSR$^\dag$} &TMI 24
  &82.45 &84.74 &85.95 &85.03 &84.54\\
  \hline
  {GIN$^\dag$} &TMI 22
  &80.56 &81.82 &83.19 &85.91 &82.87\\
  {GIN w/SDCL} &/
  &81.94 &83.51 &84.79 &86.64 &\textbf{84.22}\\
  \hline
  {SLAug$^\dag$} &AAAI 23
  &82.94 &84.73 &85.98 &85.58 &84.80\\
  {SLAug w/SDCL} &/
  &84.73 &86.05 &86.84 &88.19 &\textbf{86.45}\\
  \hline
  \end{tabular}}    
\vspace{-4mm}
\end{table}

\subsubsection{Experiment Results}

Columns 3-7 of Table~\ref{tab:SDG_Med} display the single-domain DG average Dice scores for the cross-device cardiac dataset, where our method demonstrates notable performance improvements over the four original data augmentation techniques. For instance, our approach increases the average Dice score of RandConv from 72.55\% to 73.76\% and that of UDP from 81.42\% to 83.19\%. Additionally, we observed significant enhancements in average performance across all three generalization directions. Our method significantly outperforms the others based on the SLAug technique, achieving the highest average Dice score on this task. This underscores that SDCL positively contributes to medical image segmentation. Columns 8-10 illustrate performance on the cross-modality abdominal dataset, where style differences are more pronounced. Despite the challenges, our method consistently shows performance gains across two generalization directions. The advantages of SDCL arise from causal interventions, improving the overall accuracy of RandConv, UDP, and SLAug by 2.8\%, 0.72\%, and 1.11\%, respectively, enhancing generalization performance by addressing confounding styles. Table~\ref{tab:MDG_Med} shows the average Dice scores for multi-domain generalization on the cross-device cardiac dataset. Our method improved GIN and SLAug by 1.35\% and 1.65\%, respectively, surpassing the Trid method to achieve first place. This experiment further demonstrates the versatility of SDCL.

\subsection{Alation Study}
\noindent \textbf{1) Module Effectiveness:} To investigate the contribution of the different components in SDCL, we first performed an extensive ablation study on three different tasks. Table~\ref{ablation_study} summarizes the results. \textit{Base} refers to the base method, i.e., Digits, PACS, GTAV, and CMAD correspond to UDP, AdvST, TLDR and SLAug, respectively. \textit{Base-SG} refers to adding our designed SGEM module to the \textit{Base}. As demonstrated in Table~\ref{ablation_study}, both the SGEM and BDCL modules are effective. The SGEM module employs various experts to address domain-specific styles. At the same time, BDCL effectively mitigates source-domain confounding style bias through a causal intervention mechanism, allowing for a stronger focus on domain-generalized features.

\begin{figure*}[t]
\includegraphics[width=1\textwidth]{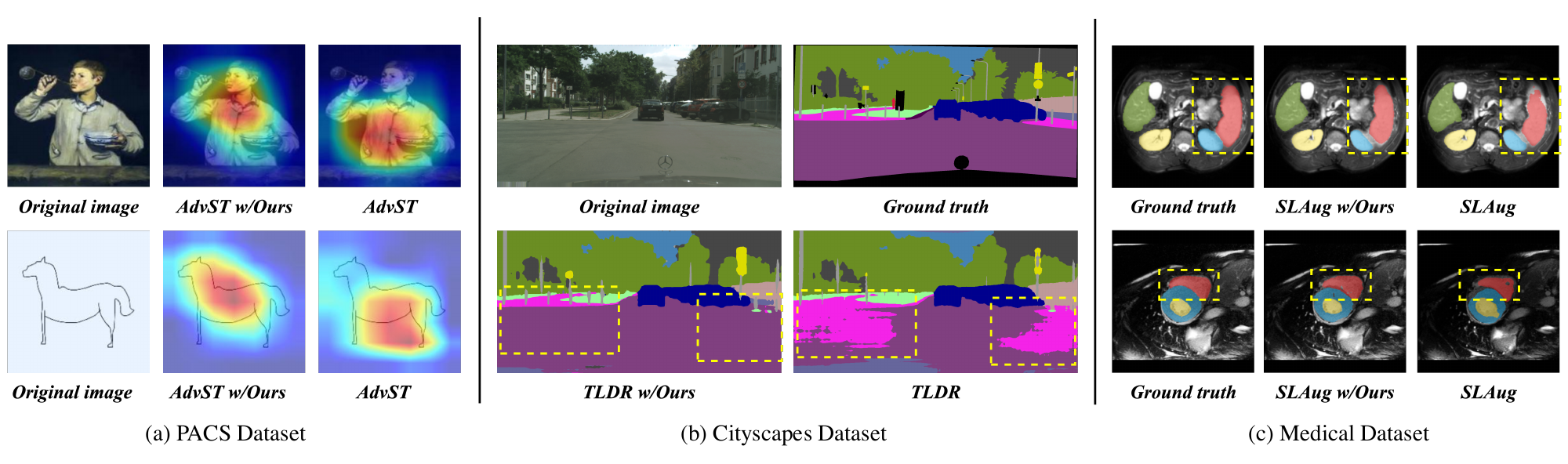}
\caption{\textbf{Visualization}. (a) Grad-CAM on \textit{PACS} dataset (''Photo'' $\rightarrow$ “Art/Sketch”). (b) Semantic segmentation example on the unseen \textit{Cityscapes} domain, using a model trained on the \textit{GTAV} dataset. (c) Semantic segmentation examples on the unseen abdominal MR domain, using a model trained on abdominal CT images, and on the unseen cardiac domain B, using a model trained on cardiac domain A.}
\label{vis}
\vspace{-2mm}
\end{figure*}

\begin{table}[t] 
    \centering
    \caption{Ablation studies of \textit{PACS, GTAV} and \textit{CMAD} under single-domain DG.}
    \label{ablation_study}
    \footnotesize
    \renewcommand\arraystretch{1.8} 
    \setlength{\tabcolsep}{1.6mm}{
    \begin{tabular}{c|cc|cccc}
    \hline
      Methods
      &SGEM &BDCL &Digits &PACS &GTAV &CMAD\\
      \hline
      {Base} &\ding{55} &\ding{55} &82.52 &68.07 &47.00 &84.39\\
      {Base-SG}   &\ding{51}  &\ding{55} &83.81 &69.83 &47.93 &85.12\\
      {SDCL} &\ding{51} &\ding{51}   &\textbf{84.36} &\textbf{71.00} &\textbf{48.47} &\textbf{85.50}\\
     \hline
    \end{tabular}}        
\end{table}

\begin{figure}[t]
\centering
\includegraphics[width=0.5 \textwidth]{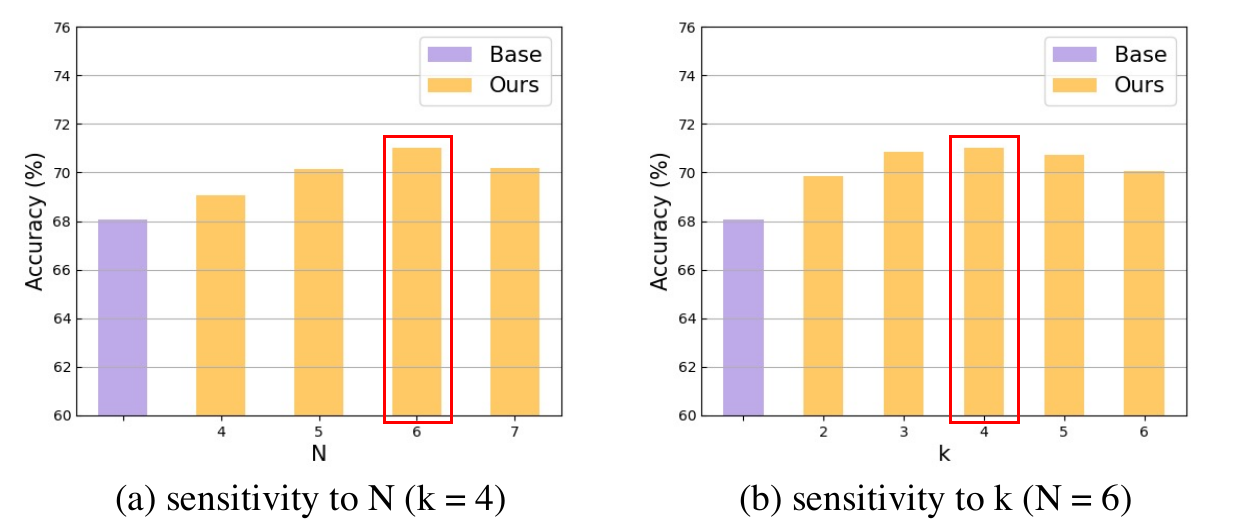}
\vspace{-4mm}
\caption{Hyper-parameter sensitivity analysis on \textit{PACS}. $N$ denotes the number of experts, and $k$ is the the TopK experts selected.}
\label{fig:hyper-k}
\vspace{-4mm}
\end{figure}

\vspace{2mm}
\noindent \textbf{2) Hyper-parameter Sensitivity:} Next, we investigated the hyperparameter sensitivity of $N$ and $k$ on the natural image classification task. $N$ is the total number of experts, and $k$ is the number of experts selected for each step. As shown in Fig.~\ref{fig:hyper-k}, the accuracy is relatively stable when each hyperparameter is varied. Because $N$ and $k$ depend on the complexity of the source domain style distribution, their values vary from one dataset to another. In our implementation, we set $N$ to 6 and $k$ to 4 for the natural image-related dataset and $N$ to 4 and $k$ to 2 for the medical image-related dataset.

\vspace{2mm}
\noindent \textbf{3) Impact of SGEM Location:} We evaluated the effect of varying the positions of multiple experts on model performance in PACS, GTAV5, and CMAD by integrating the SGEM into different layers of ResNet and the experimental results are shown in Table~\ref{layer_locations}.
The results indicate that model performance is maximized when SGEM is situated in layer 1. Performance gradually declines as the location of the experts shifts from shallow to deeper layers. This trend can be attributed to the ability of shallow layers to capture rich stylistic information, which is crucial for effectively mitigating style bias through the characterization of style distributions and causal interventions utilizing multi-domain experts.
\begin{table}[h] 
    \centering
    \caption{Performance (\%) of SGEM at different layer locations of ResNet on \textit{PACS, GTAV5}, and \textit{CMAD}.}
    \label{layer_locations}
    \footnotesize
    \renewcommand\arraystretch{1.5} 
    \setlength{\tabcolsep}{3mm}{
    \begin{tabular}{c|cccc}
    \hline
      Dataset
      &Layer 1 &Layer 2 &Layer 3 &Layer 4\\
      \hline
      {PACS} &\textbf{71.00} &70.23 &68.54 &67.12\\
      {GTAV} &\textbf{48.47}  &48.03  &46.82  &45.39\\
      {CMAD} &\textbf{85.50}  &83.04  &81.84  &80.40\\
     \hline
    \end{tabular}}        
\end{table}

\begin{figure}[t]
\centering
\includegraphics[width=0.5 \textwidth]{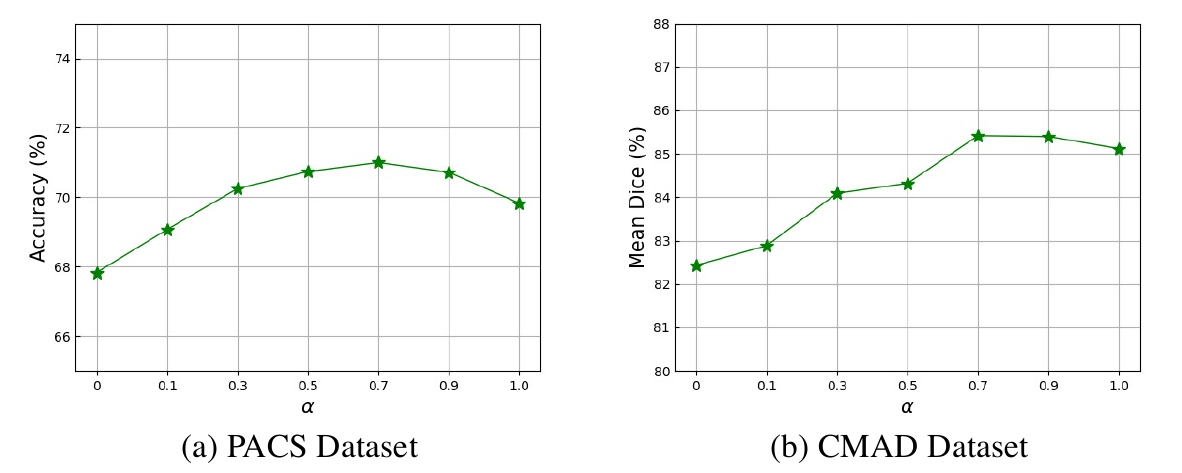}
\caption{Performance (\%) differences between different causal fusion coefficient $\alpha$ on \textit{PACS} and \textit{CMAD}.}
\label{hyper-a}
\vspace{-4mm}
\end{figure}

\noindent \textbf{4) Impact of Causal Fusion Coefficient $\mathbf{\alpha}$:} We investigated the influence of different values of $\alpha$ on model performance to effectively fuse the original feature $X$ with the confounding style feature $S$ for intervention purposes. As illustrated in Fig.~\ref{hyper-a}, when $\alpha$ is set to 0, the hybrid features consist solely of the style feature $S$ and exclude the original feature $X$, resulting in low performance. As $\alpha$ increases towards 0.7, the model performance improves, reaching its peak. Conversely, when $\alpha$ is set to 1, the hybrid feature contains only the original feature $X$, disregarding the contribution of the confounding style, which leads to a drop in performance.

\begin{figure}[h]
\includegraphics[width=0.4\textwidth, height=0.3\textwidth]{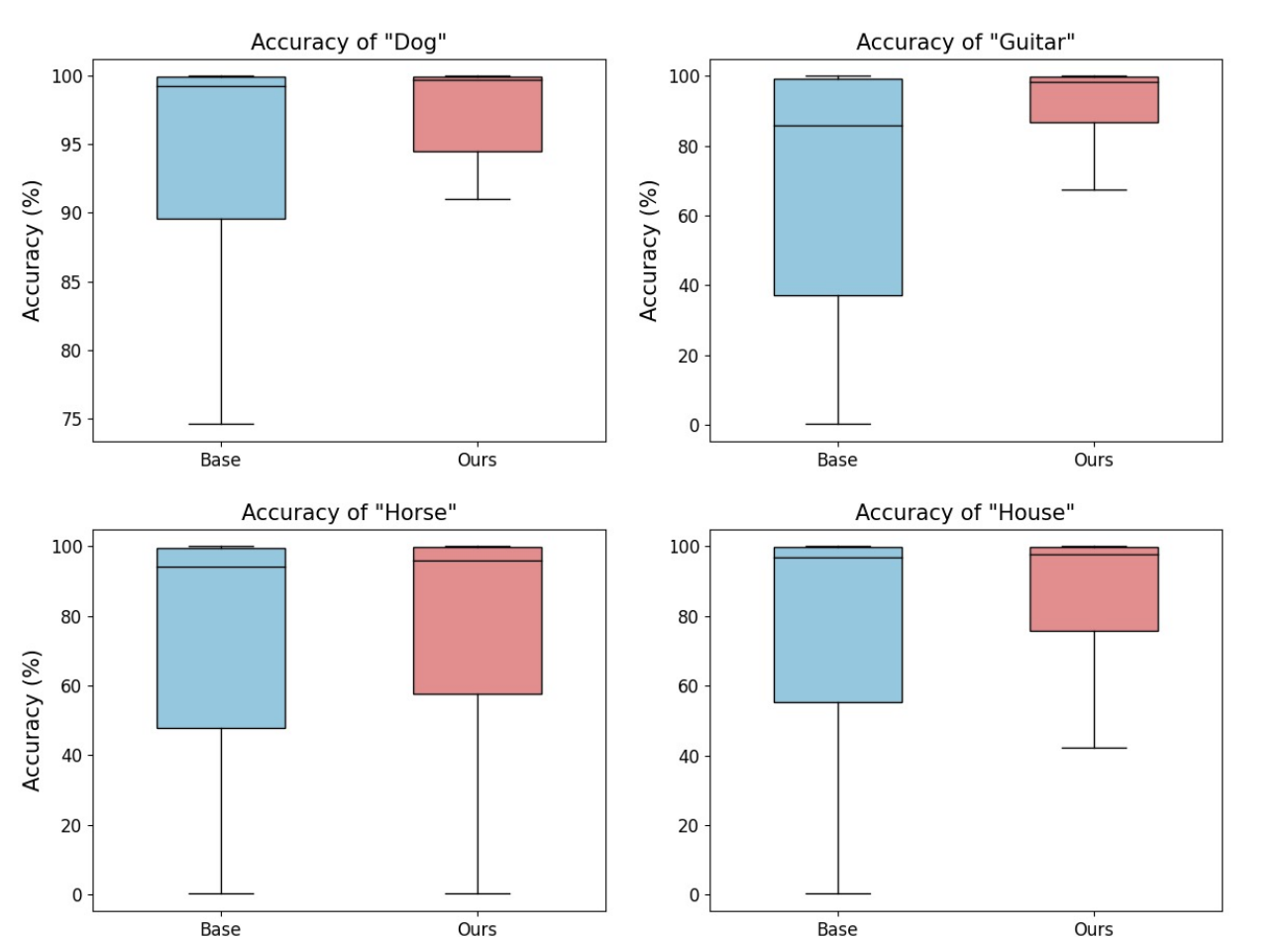}
\centering
\caption{Comparison of accuracy distributions corresponding to the four categories before and after the intervention in the \textit{PACS} dataset.}
\label{bias_effect}
\end{figure}

\vspace{2mm}
\noindent \textbf{5) Bias Elimination Effect:} Fig.~\ref{bias_effect} illustrates the distribution of model prediction accuracy before and after the intervention, with the height of the boxes indicating the degree of concentration in the data distribution. Shorter boxes represent a more concentrated distribution of predictions, highlighting reduced variability. The results demonstrate that the SDCL achieves Shorter boxes and better predictive performance across categories. This suggests that by incorporating causal intervention, SDCL significantly minimizes prediction bias, enhancing its ability to generalize to previously unseen settings.

\begin{figure}[h]
\includegraphics[width=0.5 \textwidth]{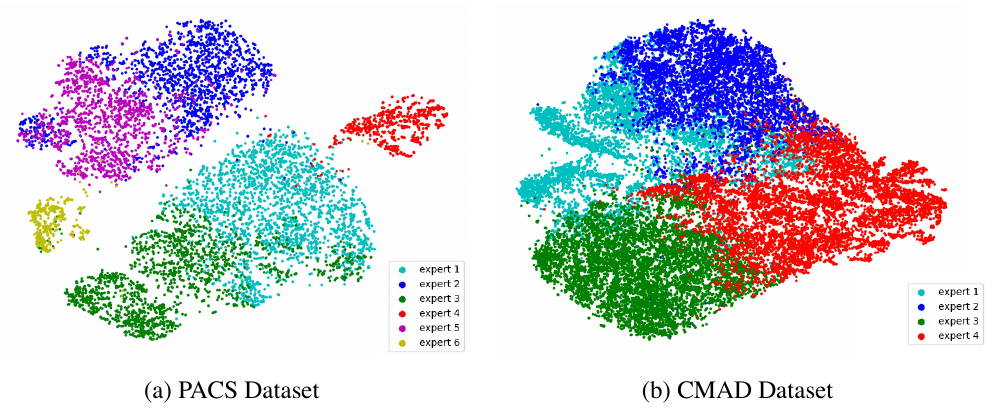}
\caption{Style distribution $t$-SNE~\cite{van2008visualizing} visualization of different experts on \textit{PACS} and \textit{CMAD} datasets.}
\label{style_vis}
\end{figure}

\vspace{2mm}
\noindent \textbf{6) Style Distribution Visualization:} We explore the style distributions selected by different experts in SGEM on the PACS and CMAD datasets. As illustrated in Fig.~\ref{style_vis}, the various colors represent different experts based on style features. SGEM adaptively clusters the style distributions of the source domains during training, forming multiple domain experts that effectively address different style distributions. This adaptability provides a solid foundation for applying hybrid style hierarchies in causal interventions.

\subsection{Visualization}
In addition to quantitative performance comparisons, we further present some qualitative illustrative results. Fig.~\ref{vis} (a) presents Grad-CAM~\cite{selvaraju2017grad} visualizations before and after applying SDCL on the PACS benchmark. The results indicate that the integration of the SDCL enhances the ability of the model to focus on critical semantic information when tested in the unseen domain, outperforming the benchmark AdvST. Fig.~\ref{vis} (b) features an example of semantic segmentation, demonstrating that SDCL effectively improves upon the TLDR baseline. The visualization highlights how SDCL corrects segmentation errors and produces more reliable predictions. Lastly, Fig.~\ref{vis} (c) displays the abdomen and cardiac segmentation results. These results confirm that SDCL significantly enhances segmentation accuracy compared to the SLAug baseline, showcasing its effectiveness in medical images.

\section{Conclusion}
This paper proposes the style deconfounding causal learning framework (SDCL) to address the domain generalization problem. The core idea is to use causal inference to effectively mitigate the impact of style bias present in the training data. To simulate causal intervention processes, we first design a style-guided expert module to cluster style distributions, construct confounded sets, and then integrate sample features from multiple style distributions based on a back-door causal learning module, facilitating causal interventions and decision-making. The effectiveness of our method is demonstrated in recognition tasks for both natural and medical images.

We provide insights for exploring causal analysis in domain generalization. In the future, we will investigate more comprehensive causal inference methods to learn better the true causal relationships between data and labels, particularly semantics. By promoting fine-grained alignment of semantic information across different distributions, we can develop a model that seeks better causal relationships. Moreover, since some confounding factors are unobservable, applying front-door adjustment to mitigate further the false associations caused by such confounding factors is also challenging. Therefore, we will explore more intuitive and rational methods to reduce spurious correlations and enhance model generalization.

\textbf{Acknowledgments.} This work is supported by the National Natural Science Foundation of China (No. 62476196)

\bibliographystyle{IEEEtran}
\bibliography{IEEEabrv,section7-bibliography}

\begin{thebibliography}{10}
\providecommand{\url}[1]{#1}
\csname url@samestyle\endcsname
\providecommand{\newblock}{\relax}
\providecommand{\bibinfo}[2]{#2}
\providecommand{\BIBentrySTDinterwordspacing}{\spaceskip=0pt\relax}
\providecommand{\BIBentryALTinterwordstretchfactor}{4}
\providecommand{\BIBentryALTinterwordspacing}{\spaceskip=\fontdimen2\font plus
\BIBentryALTinterwordstretchfactor\fontdimen3\font minus \fontdimen4\font\relax}
\providecommand{\BIBforeignlanguage}[2]{{%
\expandafter\ifx\csname l@#1\endcsname\relax
\typeout{** WARNING: IEEEtran.bst: No hyphenation pattern has been}%
\typeout{** loaded for the language `#1'. Using the pattern for}%
\typeout{** the default language instead.}%
\else
\language=\csname l@#1\endcsname
\fi
#2}}
\providecommand{\BIBdecl}{\relax}
\BIBdecl

\bibitem{ben2010theory}
S.~Ben-David, J.~Blitzer, K.~Crammer, A.~Kulesza, F.~Pereira, and J.~W. Vaughan, ``A theory of learning from different domains,'' \emph{Machine learning}, vol.~79, pp. 151--175, 2010.

\bibitem{vapnik1991principles}
V.~Vapnik, ``Principles of risk minimization for learning theory,'' \emph{Advances in neural information processing systems}, vol.~4, 1991.

\bibitem{zhu2022localized}
W.~Zhu, L.~Lu, J.~Xiao, M.~Han, J.~Luo, and A.~P. Harrison, ``Localized adversarial domain generalization,'' in \emph{Proceedings of the IEEE/CVF Conference on Computer Vision and Pattern Recognition}, 2022, pp. 7108--7118.

\bibitem{peng2019moment}
X.~Peng, Q.~Bai, X.~Xia, Z.~Huang, K.~Saenko, and B.~Wang, ``Moment matching for multi-source domain adaptation,'' in \emph{Proceedings of the IEEE/CVF international conference on computer vision}, 2019, pp. 1406--1415.

\bibitem{choi2023progressive}
S.~Choi, D.~Das, S.~Choi, S.~Yang, H.~Park, and S.~Yun, ``Progressive random convolutions for single domain generalization,'' in \emph{Proceedings of the IEEE/CVF Conference on Computer Vision and Pattern Recognition}, 2023, pp. 10\,312--10\,322.

\bibitem{cheng2023adversarial}
S.~Cheng, T.~Gokhale, and Y.~Yang, ``Adversarial bayesian augmentation for single-source domain generalization,'' in \emph{Proceedings of the IEEE/CVF International Conference on Computer Vision}, 2023, pp. 11\,400--11\,410.

\bibitem{wan2022meta}
C.~Wan, X.~Shen, Y.~Zhang, Z.~Yin, X.~Tian, F.~Gao, J.~Huang, and X.-S. Hua, ``Meta convolutional neural networks for single domain generalization,'' in \emph{Proceedings of the IEEE/CVF Conference on Computer Vision and Pattern Recognition}, 2022, pp. 4682--4691.

\bibitem{wang2022contrastive}
Y.~Wang, F.~Liu, Z.~Chen, Y.-C. Wu, J.~Hao, G.~Chen, and P.-A. Heng, ``Contrastive-ace: Domain generalization through alignment of causal mechanisms,'' \emph{IEEE Transactions on Image Processing}, vol.~32, pp. 235--250, 2022.

\bibitem{mahajan2021domain}
D.~Mahajan, S.~Tople, and A.~Sharma, ``Domain generalization using causal matching,'' in \emph{International conference on machine learning}.\hskip 1em plus 0.5em minus 0.4em\relax PMLR, 2021, pp. 7313--7324.

\bibitem{lv2022causality}
F.~Lv, J.~Liang, S.~Li, B.~Zang, C.~H. Liu, Z.~Wang, and D.~Liu, ``Causality inspired representation learning for domain generalization,'' in \emph{Proceedings of the IEEE/CVF conference on computer vision and pattern recognition}, 2022, pp. 8046--8056.

\bibitem{chen2023meta}
J.~Chen, Z.~Gao, X.~Wu, and J.~Luo, ``Meta-causal learning for single domain generalization,'' in \emph{Proceedings of the IEEE/CVF Conference on Computer Vision and Pattern Recognition}, 2023, pp. 7683--7692.

\bibitem{nguyen2023causal}
T.~Nguyen, K.~Do, D.~T. Nguyen, B.~Duong, and T.~Nguyen, ``Causal inference via style transfer for out-of-distribution generalisation,'' in \emph{Proceedings of the 29th ACM SIGKDD Conference on Knowledge Discovery and Data Mining}, 2023, pp. 1746--1757.

\bibitem{hogan2019causal}
J.~W. Hogan, ``Causal inference in statistics: A primer judea pearl, maria glymour, and nicholas jewell, john wiley \& sons, ltd., chichester, uk.'' 2019.

\bibitem{mantel1959statistical}
N.~Mantel and W.~Haenszel, ``Statistical aspects of the analysis of data from retrospective studies of disease,'' \emph{Journal of the national cancer institute}, vol.~22, no.~4, pp. 719--748, 1959.

\bibitem{guo2023single}
K.~Guo, R.~Ding, T.~Qiu, X.~Zhu, Z.~Wu, L.~Wang, and H.~Fang, ``Single domain generalization via unsupervised diversity probe,'' in \emph{Proceedings of the 31st ACM International Conference on Multimedia}, 2023, pp. 2101--2111.

\bibitem{zhao2024style}
Y.~Zhao, Z.~Zhong, N.~Zhao, N.~Sebe, and G.~H. Lee, ``Style-hallucinated dual consistency learning: A unified framework for visual domain generalization,'' \emph{International Journal of Computer Vision}, vol. 132, no.~3, pp. 837--853, 2024.

\bibitem{chattopadhyay2023pasta}
P.~Chattopadhyay, K.~Sarangmath, V.~Vijaykumar, and J.~Hoffman, ``Pasta: Proportional amplitude spectrum training augmentation for syn-to-real domain generalization,'' in \emph{Proceedings of the IEEE/CVF International Conference on Computer Vision}, 2023, pp. 19\,288--19\,300.

\bibitem{li2021progressive}
L.~Li, K.~Gao, J.~Cao, Z.~Huang, Y.~Weng, X.~Mi, Z.~Yu, X.~Li, and B.~Xia, ``Progressive domain expansion network for single domain generalization,'' in \emph{Proceedings of the IEEE/CVF Conference on Computer Vision and Pattern Recognition}, 2021, pp. 224--233.

\bibitem{pearl2010introduction}
J.~Pearl, ``An introduction to causal inference,'' \emph{The international journal of biostatistics}, vol.~6, no.~2, 2010.

\bibitem{liu2023cross}
Y.~Liu, G.~Li, and L.~Lin, ``Cross-modal causal relational reasoning for event-level visual question answering,'' \emph{IEEE Transactions on Pattern Analysis and Machine Intelligence}, vol.~45, no.~10, pp. 11\,624--11\,641, 2023.

\bibitem{zeiler2014visualizing}
M.~D. Zeiler and R.~Fergus, ``Visualizing and understanding convolutional networks,'' in \emph{Computer Vision--ECCV 2014: 13th European Conference, Zurich, Switzerland, September 6-12, 2014, Proceedings, Part I 13}.\hskip 1em plus 0.5em minus 0.4em\relax Springer, 2014, pp. 818--833.

\bibitem{shazeer2017outrageously}
N.~Shazeer, A.~Mirhoseini, K.~Maziarz, A.~Davis, Q.~Le, G.~Hinton, and J.~Dean, ``Outrageously large neural networks: The sparsely-gated mixture-of-experts layer,'' \emph{arXiv preprint arXiv:1701.06538}, 2017.

\bibitem{huang2017arbitrary}
X.~Huang and S.~Belongie, ``Arbitrary style transfer in real-time with adaptive instance normalization,'' in \emph{Proceedings of the IEEE international conference on computer vision}, 2017, pp. 1501--1510.

\bibitem{hong2021stylemix}
M.~Hong, J.~Choi, and G.~Kim, ``Stylemix: Separating content and style for enhanced data augmentation,'' in \emph{Proceedings of the IEEE/CVF conference on computer vision and pattern recognition}, 2021, pp. 14\,862--14\,870.

\bibitem{abdi2010coefficient}
H.~Abdi, ``Coefficient of variation,'' \emph{Encyclopedia of research design}, vol.~1, no.~5, pp. 169--171, 2010.

\bibitem{zhang2020causal}
D.~Zhang, H.~Zhang, J.~Tang, X.-S. Hua, and Q.~Sun, ``Causal intervention for weakly-supervised semantic segmentation,'' \emph{Advances in Neural Information Processing Systems}, vol.~33, pp. 655--666, 2020.

\bibitem{xu2015show}
K.~Xu, J.~Ba, R.~Kiros, K.~Cho, A.~Courville, R.~Salakhudinov, R.~Zemel, and Y.~Bengio, ``Show, attend and tell: Neural image caption generation with visual attention,'' in \emph{International conference on machine learning}.\hskip 1em plus 0.5em minus 0.4em\relax PMLR, 2015, pp. 2048--2057.

\bibitem{vapnik1999overview}
V.~N. Vapnik, ``An overview of statistical learning theory,'' \emph{IEEE transactions on neural networks}, vol.~10, no.~5, pp. 988--999, 1999.

\bibitem{addepalli2024leveraging}
S.~Addepalli, A.~R. Asokan, L.~Sharma, and R.~V. Babu, ``Leveraging vision-language models for improving domain generalization in image classification,'' in \emph{Proceedings of the IEEE/CVF Conference on Computer Vision and Pattern Recognition}, 2024, pp. 23\,922--23\,932.

\bibitem{ding2022domain}
Y.~Ding, L.~Wang, B.~Liang, S.~Liang, Y.~Wang, and F.~Chen, ``Domain generalization by learning and removing domain-specific features,'' \emph{Advances in Neural Information Processing Systems}, vol.~35, pp. 24\,226--24\,239, 2022.

\bibitem{huang2020self}
Z.~Huang, H.~Wang, E.~P. Xing, and D.~Huang, ``Self-challenging improves cross-domain generalization,'' in \emph{Computer vision--ECCV 2020: 16th European conference, Glasgow, UK, August 23--28, 2020, proceedings, part II 16}.\hskip 1em plus 0.5em minus 0.4em\relax Springer, 2020, pp. 124--140.

\bibitem{L2D}
Z.~Wang, Y.~Luo, R.~Qiu, Z.~Huang, and M.~Baktashmotlagh, ``Learning to diversify for single domain generalization,'' in \emph{Proceedings of the IEEE/CVF International Conference on Computer Vision}, 2021, pp. 834--843.

\bibitem{liu2022geometric}
X.-C. Liu, Y.-L. Yang, and P.~Hall, ``Geometric and textural augmentation for domain gap reduction,'' in \emph{Proceedings of the IEEE/CVF Conference on Computer Vision and Pattern Recognition}, 2022, pp. 14\,340--14\,350.

\bibitem{lee2022cross}
K.~Lee, S.~Kim, and S.~Kwak, ``Cross-domain ensemble distillation for domain generalization,'' in \emph{European Conference on Computer Vision}.\hskip 1em plus 0.5em minus 0.4em\relax Springer, 2022, pp. 1--20.

\bibitem{qi2024normaug}
L.~Qi, H.~Yang, Y.~Shi, and X.~Geng, ``Normaug: Normalization-guided augmentation for domain generalization,'' \emph{IEEE Transactions on Image Processing}, vol.~33, pp. 1419--1431, 2024.

\bibitem{zheng2024advst}
G.~Zheng, M.~Huai, and A.~Zhang, ``Advst: Revisiting data augmentations for single domain generalization,'' in \emph{Proceedings of the AAAI Conference on Artificial Intelligence}, vol.~38, no.~19, 2024, pp. 21\,832--21\,840.

\bibitem{MNIST}
Y.~LeCun, L.~Bottou, Y.~Bengio, and P.~Haffner, ``Gradient-based learning applied to document recognition,'' \emph{Proceedings of the IEEE}, vol.~86, no.~11, pp. 2278--2324, 1998.

\bibitem{SVHN}
Y.~Netzer, T.~Wang, A.~Coates, A.~Bissacco, B.~Wu, and A.~Y. Ng, ``Reading digits in natural images with unsupervised feature learning,'' 2011.

\bibitem{MNISTM-SYN}
Y.~Ganin and V.~Lempitsky, ``Unsupervised domain adaptation by backpropagation,'' in \emph{International conference on machine learning}.\hskip 1em plus 0.5em minus 0.4em\relax PMLR, 2015, pp. 1180--1189.

\bibitem{USPS}
J.~Denker, W.~Gardner, H.~Graf, D.~Henderson, R.~Howard, W.~Hubbard, L.~D. Jackel, H.~Baird, and I.~Guyon, ``Neural network recognizer for hand-written zip code digits,'' \emph{Advances in neural information processing systems}, vol.~1, 1988.

\bibitem{li2017deeper}
D.~Li, Y.~Yang, Y.-Z. Song, and T.~M. Hospedales, ``Deeper, broader and artier domain generalization,'' in \emph{Proceedings of the IEEE international conference on computer vision}, 2017, pp. 5542--5550.

\bibitem{he2016deep}
K.~He, X.~Zhang, S.~Ren, and J.~Sun, ``Deep residual learning for image recognition,'' in \emph{Proceedings of the IEEE conference on computer vision and pattern recognition}, 2016, pp. 770--778.

\bibitem{kingma2014adam}
D.~P. Kingma and J.~Ba, ``Adam: A method for stochastic optimization,'' \emph{arXiv preprint arXiv:1412.6980}, 2014.

\bibitem{pan2018two}
X.~Pan, P.~Luo, J.~Shi, and X.~Tang, ``Two at once: Enhancing learning and generalization capacities via ibn-net,'' in \emph{Proceedings of the european conference on computer vision (ECCV)}, 2018, pp. 464--479.

\bibitem{choi2021robustnet}
S.~Choi, S.~Jung, H.~Yun, J.~T. Kim, S.~Kim, and J.~Choo, ``Robustnet: Improving domain generalization in urban-scene segmentation via instance selective whitening,'' in \emph{Proceedings of the IEEE/CVF conference on computer vision and pattern recognition}, 2021, pp. 11\,580--11\,590.

\bibitem{lee2022wildnet}
S.~Lee, H.~Seong, S.~Lee, and E.~Kim, ``Wildnet: Learning domain generalized semantic segmentation from the wild,'' in \emph{Proceedings of the IEEE/CVF conference on computer vision and pattern recognition}, 2022, pp. 9936--9946.

\bibitem{peng2022semantic}
D.~Peng, Y.~Lei, M.~Hayat, Y.~Guo, and W.~Li, ``Semantic-aware domain generalized segmentation,'' in \emph{Proceedings of the IEEE/CVF conference on computer vision and pattern recognition}, 2022, pp. 2594--2605.

\bibitem{li2018learning}
D.~Li, Y.~Yang, Y.-Z. Song, and T.~Hospedales, ``Learning to generalize: Meta-learning for domain generalization,'' in \emph{Proceedings of the AAAI conference on artificial intelligence}, vol.~32, no.~1, 2018.

\bibitem{zhong2022adversarial}
Z.~Zhong, Y.~Zhao, G.~H. Lee, and N.~Sebe, ``Adversarial style augmentation for domain generalized urban-scene segmentation,'' \emph{Advances in neural information processing systems}, vol.~35, pp. 338--350, 2022.

\bibitem{kim2023texture}
S.~Kim, D.-h. Kim, and H.~Kim, ``Texture learning domain randomization for domain generalized segmentation,'' in \emph{Proceedings of the IEEE/CVF International Conference on Computer Vision}, 2023, pp. 677--687.

\bibitem{richter2016playing}
S.~R. Richter, V.~Vineet, S.~Roth, and V.~Koltun, ``Playing for data: Ground truth from computer games,'' in \emph{Computer Vision--ECCV 2016: 14th European Conference, Amsterdam, The Netherlands, October 11-14, 2016, Proceedings, Part II 14}.\hskip 1em plus 0.5em minus 0.4em\relax Springer, 2016, pp. 102--118.

\bibitem{ros2016synthia}
G.~Ros, L.~Sellart, J.~Materzynska, D.~Vazquez, and A.~M. Lopez, ``The synthia dataset: A large collection of synthetic images for semantic segmentation of urban scenes,'' in \emph{Proceedings of the IEEE conference on computer vision and pattern recognition}, 2016, pp. 3234--3243.

\bibitem{cordts2016cityscapes}
M.~Cordts, M.~Omran, S.~Ramos, T.~Rehfeld, M.~Enzweiler, R.~Benenson, U.~Franke, S.~Roth, and B.~Schiele, ``The cityscapes dataset for semantic urban scene understanding,'' in \emph{Proceedings of the IEEE conference on computer vision and pattern recognition}, 2016, pp. 3213--3223.

\bibitem{yu2020bdd100k}
F.~Yu, H.~Chen, X.~Wang, W.~Xian, Y.~Chen, F.~Liu, V.~Madhavan, and T.~Darrell, ``Bdd100k: A diverse driving dataset for heterogeneous multitask learning,'' in \emph{Proceedings of the IEEE/CVF conference on computer vision and pattern recognition}, 2020, pp. 2636--2645.

\bibitem{neuhold2017mapillary}
G.~Neuhold, T.~Ollmann, S.~Rota~Bulo, and P.~Kontschieder, ``The mapillary vistas dataset for semantic understanding of street scenes,'' in \emph{Proceedings of the IEEE international conference on computer vision}, 2017, pp. 4990--4999.

\bibitem{chen2018encoder}
L.-C. Chen, Y.~Zhu, G.~Papandreou, F.~Schroff, and H.~Adam, ``Encoder-decoder with atrous separable convolution for semantic image segmentation,'' in \emph{Proceedings of the European conference on computer vision (ECCV)}, 2018, pp. 801--818.

\bibitem{liu2022vmfnet}
X.~Liu, S.~Thermos, P.~Sanchez, A.~Q. O’Neil, and S.~A. Tsaftaris, ``vmfnet: Compositionality meets domain-generalised segmentation,'' in \emph{International Conference on Medical Image Computing and Computer-Assisted Intervention}.\hskip 1em plus 0.5em minus 0.4em\relax Springer, 2022, pp. 704--714.

\bibitem{gu2023cddsa}
R.~Gu, G.~Wang, J.~Lu, J.~Zhang, W.~Lei, Y.~Chen, W.~Liao, S.~Zhang, K.~Li, D.~N. Metaxas \emph{et~al.}, ``Cddsa: Contrastive domain disentanglement and style augmentation for generalizable medical image segmentation,'' \emph{Medical Image Analysis}, vol.~89, p. 102904, 2023.

\bibitem{chen2024learning}
K.~Chen, T.~Qin, V.~H.-F. Lee, H.~Yan, and H.~Li, ``Learning robust shape regularization for generalizable medical image segmentation,'' \emph{IEEE Transactions on Medical Imaging}, 2024.

\bibitem{chen2022maxstyle}
C.~Chen, Z.~Li, C.~Ouyang, M.~Sinclair, W.~Bai, and D.~Rueckert, ``Maxstyle: Adversarial style composition for robust medical image segmentation,'' in \emph{International Conference on Medical Image Computing and Computer-Assisted Intervention}.\hskip 1em plus 0.5em minus 0.4em\relax Springer, 2022, pp. 151--161.

\bibitem{chen2023treasure}
Z.~Chen, Y.~Pan, Y.~Ye, H.~Cui, and Y.~Xia, ``Treasure in distribution: a domain randomization based multi-source domain generalization for 2d medical image segmentation,'' in \emph{International Conference on Medical Image Computing and Computer-Assisted Intervention}.\hskip 1em plus 0.5em minus 0.4em\relax Springer, 2023, pp. 89--99.

\bibitem{ouyang2022causality}
C.~Ouyang, C.~Chen, S.~Li, Z.~Li, C.~Qin, W.~Bai, and D.~Rueckert, ``Causality-inspired single-source domain generalization for medical image segmentation,'' \emph{IEEE Transactions on Medical Imaging}, vol.~42, no.~4, pp. 1095--1106, 2022.

\bibitem{su2023rethinking}
Z.~Su, K.~Yao, X.~Yang, K.~Huang, Q.~Wang, and J.~Sun, ``Rethinking data augmentation for single-source domain generalization in medical image segmentation,'' in \emph{Proceedings of the AAAI Conference on Artificial Intelligence}, vol.~37, no.~2, 2023, pp. 2366--2374.

\bibitem{xurobust}
Z.~Xu, D.~Liu, J.~Yang, C.~Raffel, and M.~Niethammer, ``Robust and generalizable visual representation learning via random convolutions,'' in \emph{International Conference on Learning Representations}, 2021.

\bibitem{kavur2021chaos}
A.~E. Kavur, N.~S. Gezer, M.~Bar{\i}{\c{s}}, S.~Aslan, P.-H. Conze, V.~Groza, D.~D. Pham, S.~Chatterjee, P.~Ernst, S.~{\"O}zkan \emph{et~al.}, ``Chaos challenge-combined (ct-mr) healthy abdominal organ segmentation,'' \emph{Medical Image Analysis}, vol.~69, p. 101950, 2021.

\bibitem{landman2015miccai}
B.~Landman, Z.~Xu, J.~Igelsias, M.~Styner, T.~Langerak, and A.~Klein, ``Miccai multi-atlas labeling beyond the cranial vault--workshop and challenge,'' in \emph{Proc. MICCAI Multi-Atlas Labeling Beyond Cranial Vault—Workshop Challenge}, vol.~5, 2015, p.~12.

\bibitem{campello2021multi}
V.~M. Campello, P.~Gkontra, C.~Izquierdo, C.~Martin-Isla, A.~Sojoudi, P.~M. Full, K.~Maier-Hein, Y.~Zhang, Z.~He, J.~Ma \emph{et~al.}, ``Multi-centre, multi-vendor and multi-disease cardiac segmentation: the m\&ms challenge,'' \emph{IEEE Transactions on Medical Imaging}, vol.~40, no.~12, pp. 3543--3554, 2021.

\bibitem{van2008visualizing}
L.~Van~der Maaten and G.~Hinton, ``Visualizing data using t-sne.'' \emph{Journal of machine learning research}, vol.~9, no.~11, 2008.

\bibitem{selvaraju2017grad}
R.~R. Selvaraju, M.~Cogswell, A.~Das, R.~Vedantam, D.~Parikh, and D.~Batra, ``Grad-cam: Visual explanations from deep networks via gradient-based localization,'' in \emph{Proceedings of the IEEE international conference on computer vision}, 2017, pp. 618--626.

\end{thebibliography}

\end{document}